\documentclass[draft]{agujournal2019}
\usepackage{url} 
\usepackage{lineno}
\usepackage[inline]{trackchanges} 
\usepackage{soul}
\usepackage{booktabs}
\justifying
\draftfalse

\journalname{arXiv}

\begin{document}

%
%


\title{
Physical Sensitivity Kernels Can Emerge in Data-Driven Forward Models: Evidence From Surface-Wave Dispersion
}

%
%




\authors{
        Ziye Yu\affil{1}, 
        Yuqi Cai\affil{1}, 
        Xin Liu\affil{2,3}, 
        }

\affiliation{1}{Institute of Geophysics, China Earthquake Administration,Beijing 100081, China}
\affiliation{2}{Laboratory of Seismology and Physics of Earth’s Interior, School of Earth and Space Sciences, University of Science and Technology of China, Hefei 230026, China}
\affiliation{3}{Institute of Advanced Technology, University of Science and Technology of China, Hefei 230088, China}




\correspondingauthor{Ziye Yu}{yuziye@cea-igp.ac.cn}



\begin{keypoints}
\item Neural network surrogates recover the main depth-dependent structure of surface-wave sensitivity kernels.
\item Learned sensitivities are shaped by both wave physics and the training distribution.
\item Surrogate gradients and Fisher information capture useful local inverse-problem geometry for inversion.
\end{keypoints}

%
%

%
%


\begin{abstract}
Data-driven neural networks are increasingly used as surrogate forward models in geophysics, but it remains unclear whether they recover only the data mapping or also the underlying physical sensitivity structure. Here we test this question using surface-wave dispersion. By comparing automatically differentiated gradients from a neural-network surrogate with theoretical sensitivity kernels, we show that the learned gradients can recover the main depth-dependent structure of physical kernels across a broad range of periods. This indicates that neural surrogate models can learn physically meaningful differential information, rather than acting as purely black-box predictors. At the same time, strong structural priors in the training distribution can introduce systematic artifacts into the inferred sensitivities. Our results show that neural forward surrogates can recover useful physical information for inversion and uncertainty analysis, while clarifying the conditions under which this differential structure remains physically consistent.
\end{abstract}

\section*{Plain Language Summary}
Machine learning models are increasingly used in geophysics to approximate complex physical calculations that would otherwise be computationally expensive. However, it remains unclear whether these models capture the underlying physical behavior of the Earth system or merely reproduce observations. In this study, we examine this question using surface waves, which are commonly used to study the Earth's interior. We train a neural network to predict how surface‐wave speeds depend on subsurface velocity structure. By analyzing how the model’s predictions change when the subsurface structure is slightly perturbed, we obtain sensitivity patterns that describe which depths influence the observations. We find that these learned sensitivity patterns closely match those predicted by established physical theory. Furthermore, they can be used to recover subsurface velocity models from observations. These results suggest that machine learning models trained only on observable data can still recover physically meaningful information about the Earth.

%
%

%


%
%
%
%

\section{Introduction}

Data‐driven models, particularly deep neural networks, are increasingly used in geophysical research to approximate complex forward calculations and accelerate inverse problems. Applications include seismic waveform modeling, travel‐time prediction, and surface‐wave dispersion analysis, where neural networks have been trained as surrogate models to emulate computationally expensive physical simulations \cite{smith2021seismic, waheed2021efficient, zhu2023deeplearning}. By learning the mapping between model parameters and observable quantities directly from simulated or observed data, such models can significantly reduce computational cost and enable rapid exploration of parameter spaces.

Despite these advantages, an important conceptual question remains unresolved: to what extent do data‐driven models capture the underlying physical structure of the forward operator they approximate? In many geophysical applications, neural networks are trained solely to reproduce observable quantities, such as travel times or dispersion curves. While prediction accuracy can often be evaluated straightforwardly, it remains unclear whether the internal structure of the learned mapping preserves physically meaningful relationships between model parameters and observations. This issue is closely related to the widely discussed ``black box'' nature of deep learning models, which raises concerns about their interpretability and physical reliability in scientific applications.

In geophysical inversion theory, the differential structure of the forward operator plays a central role. The sensitivity of observations to model perturbations is described by Fréchet derivatives or sensitivity kernels, which provide the physical basis for gradient‐based inversion and uncertainty analysis \cite{tarantola2005inverse, dahlen1998theoretical}. For example, in surface‐wave tomography, the relationship between shear‐wave velocity structure and dispersion measurements is governed by well‐established sensitivity kernels derived from perturbation theory. These kernels quantify how variations in velocity at different depths affect phase or group velocities at different periods and form the foundation of many inversion algorithms.

An open question is whether such differential physical structures can also emerge in purely data‐driven models trained on observable quantities. If a neural network surrogate accurately approximates a forward operator, its derivatives with respect to model parameters can be computed efficiently using automatic differentiation. However, it is not obvious whether these derivatives correspond to physically meaningful sensitivity kernels or merely reflect numerical properties of the learned function. Demonstrating that the gradients of a data‐driven model reproduce the theoretical sensitivity structure would provide strong evidence that the model has learned not only the mapping between inputs and outputs, but also key aspects of the underlying physical operator.

In this study, we investigate this question using the forward problem of surface‐wave dispersion. We train neural network surrogate models to approximate the mapping from one‐dimensional shear‐wave velocity structures to Rayleigh‐wave dispersion curves across multiple periods. Using automatic differentiation, we compute the gradient of the predicted dispersion with respect to the velocity structure and compare the resulting sensitivity patterns with theoretical surface‐wave sensitivity kernels. Furthermore, we examine whether these gradients can be used directly for gradient‐based inversion, providing an additional test of their physical consistency.

Our results show that the gradients of the trained neural network closely reproduce the depth‐dependent structure of theoretical surface‐wave sensitivity kernels. Moreover, gradient‐based inversion using these automatically differentiated sensitivities successfully recovers velocity models from synthetic dispersion data. These results indicate that the differential structure of the surface‐wave dispersion operator can emerge naturally in data‐driven models trained only on observable quantities.

We refer to this phenomenon as \emph{emergent differential physics}, in which the differential physical structure of a forward operator is recovered through differentiation of a data‐driven model. The findings suggest that neural network surrogates of geophysical forward models may be more physically informative than often assumed, and that automatic differentiation of such models can provide physically consistent sensitivities for inversion and uncertainty analysis.

\section{Neural Network Surrogate for Surface-Wave Dispersion}

\subsection{Synthetic Training Data}

To construct a data‐driven surrogate forward model for surface‐wave dispersion, we generate a large ensemble of synthetic one‐dimensional Earth models and corresponding dispersion curves. Each model consists of depth‐dependent profiles of $V_p$, $V_s$, and density extending to 150 km depth. Velocity structures are sampled within physically plausible ranges to ensure diverse crust–upper mantle configurations.

For each model, Rayleigh‐ and Love‐wave phase velocities are computed over periods from 2 to 60 s using a conventional normal‐mode solver. These dispersion curves serve as reference observations for training. To mimic realistic data conditions, partial dispersion curves are generated by randomly masking period ranges and, in some cases, removing either Rayleigh or Love wave branches.

Each training sample therefore consists of a velocity model $m(z)$ and corresponding dispersion observations $d(T)$.

\subsection{Neural Network Surrogate Model}

We train a neural network to approximate the forward mapping between velocity structure and dispersion observations,

\[
f_\theta(m) \approx d,
\]

where $m$ denotes the discretized velocity model and $d$ denotes the dispersion curve. The model is designed to capture the depth‐dependent relationship between subsurface structure and surface‐wave propagation across a wide range of periods.

The network takes the depth‐dependent physical parameters as input and predicts Rayleigh‐ and Love‐wave phase velocities at multiple periods. Details of the network architecture are not critical for the present study, as our focus is on the physical consistency of the learned mapping rather than architectural design.

\subsection{Automatic Differentiation of Sensitivity Structure}

A key advantage of the surrogate model is that it provides a fully differentiable approximation of the forward operator. Gradients of predicted dispersion with respect to the velocity structure can therefore be computed using automatic differentiation,

\[
\frac{\partial d(T)}{\partial m(z)} \approx
\frac{\partial f_\theta(m)}{\partial m}.
\]

These gradients quantify the sensitivity of dispersion observations to perturbations in subsurface structure. In classical surface‐wave theory, such sensitivities are described by Fréchet kernels derived from perturbation analysis. Here we interpret the neural network gradients as data‐driven estimates of these sensitivity kernels and directly compare them with their theoretical counterparts.

In the following sections, we evaluate whether the gradients recovered from the neural surrogate reproduce the depth‐dependent sensitivity structure predicted by theory and assess their applicability in gradient‐based inversion.

\section{Results}

\subsection{Dispersion Prediction Accuracy}

The neural surrogate achieves high accuracy in predicting surface-wave dispersion across the full period range (2–60 s) for both Rayleigh and Love waves. Mean absolute percentage errors (MAPE) are generally below 1\%, with typical values ranging from $\sim$0.3\% to 1.2\% depending on period and wave type.

For Rayleigh waves, MAPE is approximately 0.9–1.1\% at short to intermediate periods (2–15 s), decreases to $\sim$0.3–0.5\% at intermediate periods (20–40 s), and increases slightly at the longest periods (50–60 s). Love waves exhibit a similar trend, with slightly lower errors at short periods and modest degradation at the longest periods. The increase in error at long periods likely reflects reduced sensitivity and increased smoothness of dispersion curves.

Representative results are summarized in Table~\ref{tab:combined_results}. Full results for all period bands are provided in the Supporting Information.

\begin{table}[!htbp]
\centering
\caption{Statistical summary of dispersion prediction accuracy and sensitivity kernel similarity.}
\label{tab:combined_results}
\small
\setlength{\tabcolsep}{4pt} 
\begin{tabular}{lccccccr}
\toprule
 & & \multicolumn{2}{c}{Dispersion Accuracy} & \multicolumn{2}{c}{Kernel Similarity} & \\
\cmidrule(lr){3-4} \cmidrule(lr){5-6}
Wave & Period (s) & MAE & MAPE (\%) & Cosine & Correlation & $N$ \\
\midrule
Rayleigh & 2–5   & 0.00963 & 0.605 & 0.9776 & 0.9771 & 3,000 \\
         & 5–10  & 0.01212 & 0.727 & 0.9820 & 0.9811 & 5,000 \\
         & 10–20 & 0.01301 & 0.778 & 0.9675 & 0.9611 & 10,000 \\
         & 20–30 & 0.01344 & 0.651 & 0.9329 & 0.9041 & 10,000 \\
         & 30–45 & 0.01401 & 0.527 & 0.8883 & 0.7889 & 15,000 \\
         & 45–60 & 0.01168 & 0.339 & 0.8246 & 0.6080 & 16,000 \\
\midrule
Love     & 2–5   & 0.01078 & 0.667 & 0.9762 & 0.9769 & 3,000 \\
         & 5–10  & 0.01024 & 0.558 & 0.9822 & 0.9805 & 5,000 \\
         & 10–20 & 0.01110 & 0.622 & 0.9763 & 0.9711 & 10,000 \\
         & 20–30 & 0.01507 & 0.840 & 0.9665 & 0.9515 & 10,000 \\
         & 30–45 & 0.01714 & 0.726 & 0.9443 & 0.9042 & 15,000 \\
         & 45–60 & 0.01446 & 0.417 & 0.9053 & 0.7878 & 16,000 \\
\bottomrule
\end{tabular}
\end{table}

These results demonstrate that the neural network provides an accurate surrogate of the dispersion forward operator, with prediction errors remaining within or below typical observational uncertainties for surface-wave dispersion measurements.

We next evaluate whether gradients obtained from automatic differentiation reproduce theoretical surface-wave sensitivity kernels. Similarity between neural gradients and theoretical kernels is quantified using cosine similarity and Pearson correlation.

At short periods (2–15 s), neural gradients show excellent agreement with theoretical kernels for both Rayleigh and Love waves, with cosine similarity exceeding 0.97 and correlation coefficients above 0.96. This indicates that the surrogate model accurately captures the shallow sensitivity structure of surface waves.

At intermediate periods (20–30 s), agreement remains high, with cosine similarity around 0.95–0.96 for both wave types. Although minor discrepancies in amplitude and smoothness begin to appear, the overall depth-dependent sensitivity structure is well preserved.

At longer periods (30–60 s), kernel agreement gradually decreases, particularly for Rayleigh waves. Cosine similarity decreases from $\sim$0.93 at 30–40 s to $\sim$0.81 at 50–60 s, with a corresponding reduction in correlation. Love-wave kernels remain comparatively more stable, maintaining cosine similarity above 0.89 even at the longest periods. Despite this degradation, the neural gradients continue to capture the broad sensitivity distribution and overall depth trends of the theoretical kernels.

Representative results are summarized in Table~\ref{tab:combined_results}. In addition, the absolute errors between neural and theoretical kernels remain small (MAE $\sim$0.001--0.003), indicating that the discrepancies are primarily structural rather than amplitude-dominated.

\subsection{Depth-Dependent Structure of Sensitivity Kernels}

\begin{figure}
    \centering
    \includegraphics[width=0.8\linewidth]{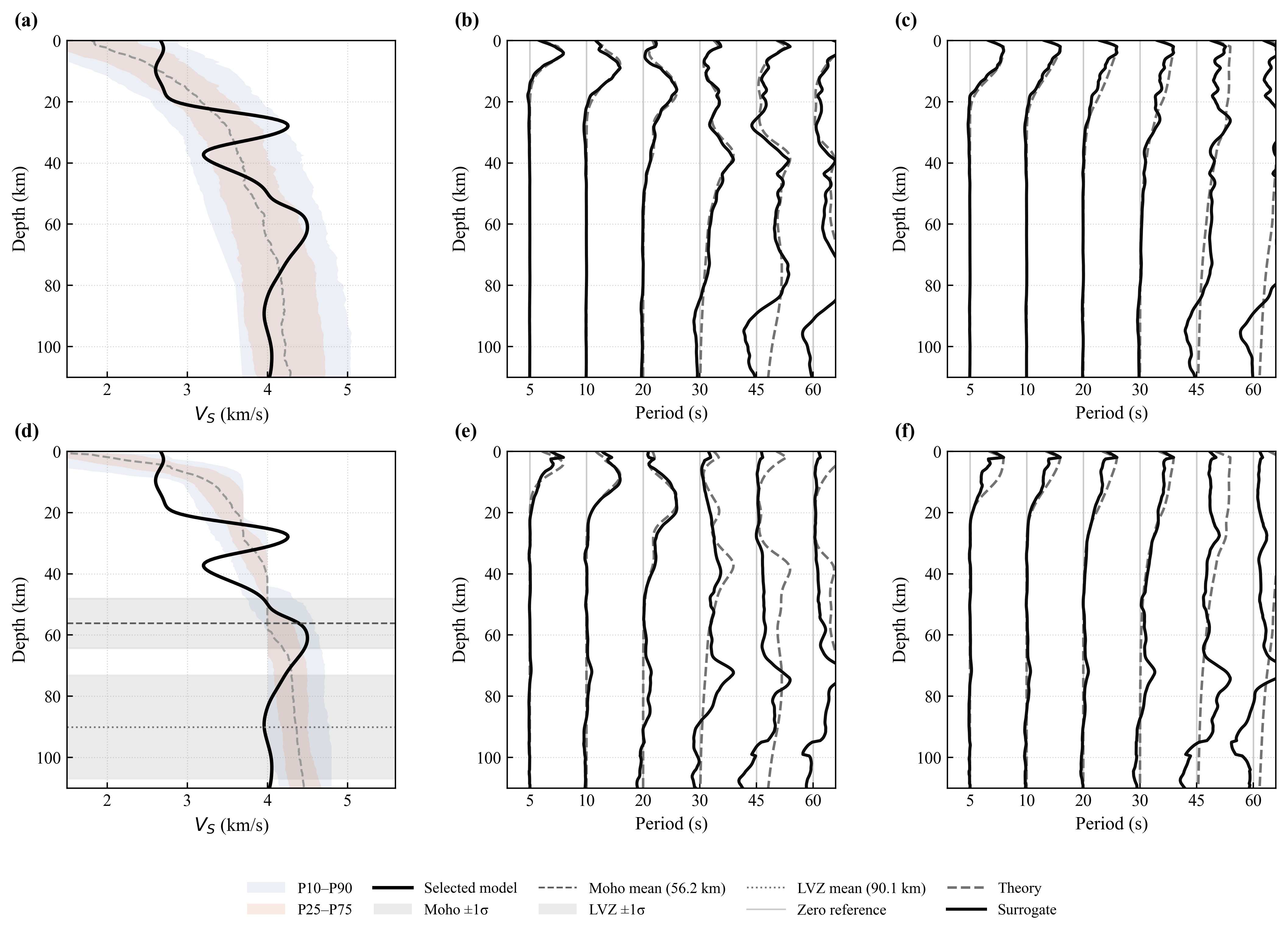}
    \caption{
Influence of training priors on surrogate-derived sensitivity kernels for a representative velocity model.
(a) Weak-prior ensemble, in which velocity models are broadly distributed without imposing specific structural features. The representative model used for all kernel calculations is shown by the black curve; shading indicates the P10–P90 and P25–P75 ranges.
(b)–(c) Rayleigh- and Love-wave sensitivity kernels computed from the surrogate trained on the weak-prior ensemble, compared with theoretical kernels. The surrogate gradients closely follow the theoretical kernels, with peak sensitivity shifting systematically with period, consistent with physical expectations.
(d) Strong-LVZ-prior ensemble, in which a low-velocity zone (LVZ) is imposed in the upper mantle. Horizontal bands and lines indicate the mean and ±1$\sigma$ ranges of the Moho depth and LVZ center inferred from the prior.
(e)–(f) Rayleigh- and Love-wave sensitivity kernels from the surrogate trained on the strong-prior ensemble, using the same representative model as in (a). In contrast to (b)–(c), the surrogate gradients exhibit persistent depth-localized anomalies near the imposed LVZ depth that remain approximately fixed across periods, deviating from theoretical kernels whose sensitivity shifts with period.
This comparison demonstrates that the physical consistency of surrogate-derived gradients depends on the training prior, with strong structural priors introducing non-physical features in the inferred sensitivity kernels.
}
    \label{fig:kernel}
\end{figure}
Figure~\ref{fig:kernel} illustrates the influence of training priors on the inferred sensitivity structure using a representative velocity model. Panels (a)–(c) show results from the weak-prior ensemble, while panels (d)–(f) correspond to the strong-LVZ-prior ensemble. Importantly, the same representative model is used in both cases, allowing a direct comparison of the effect of training distribution on the learned gradients.

In the weak-prior ensemble (Figure~\ref{fig:kernel}a), the distribution of velocity models is broad and does not enforce specific structural features. The corresponding sensitivity kernels derived from the neural surrogate (Figures~\ref{fig:kernel}b–c) show strong agreement with theoretical kernels across all periods. The depth of maximum sensitivity shifts systematically with period for both Rayleigh and Love waves, and the overall shape and amplitude of the kernels are well reproduced. This indicates that, in the absence of strong structural constraints, the neural surrogate recovers the physically expected sensitivity structure.

In contrast, the strong-prior ensemble (Figure~\ref{fig:kernel}d) imposes a systematic low-velocity zone (LVZ) in the upper mantle, as indicated by the narrow distribution of LVZ depth and Moho depth. Despite using the same input model, the sensitivity kernels derived from the surrogate trained on this ensemble (Figures~\ref{fig:kernel}e–f) exhibit pronounced deviations from theoretical kernels. In particular, a persistent depth-localized anomaly appears near the LVZ depth range, and this feature remains approximately fixed with respect to depth across different periods. This behavior contrasts with the theoretical kernels, whose peak sensitivity shifts with period, and indicates a non-physical contribution to the learned gradients.

The comparison demonstrates that the inferred sensitivity structure is not solely determined by the forward operator and the input model, but is strongly influenced by the statistical properties of the training ensemble. While the weak-prior model yields physically consistent kernels, the strong-prior model introduces systematic, depth-localized features that reflect the imposed structural prior rather than intrinsic wave physics.

\section{Discussion}

\subsection{Emergence of Differential Physical Structure in Data-Driven Models}

Our results show that neural network surrogates trained solely on input–output mappings of dispersion curves can recover the dominant depth-dependent sensitivity structure of surface waves. In particular, at short and intermediate periods, the gradients obtained through automatic differentiation closely match theoretical Fréchet kernels in both shape and depth localization. 

This behavior suggests that the differential structure of the forward operator is implicitly encoded in the training data and can emerge in the learned model without explicit physical constraints. In this sense, the neural surrogate acts as a data-driven approximation not only of the forward mapping $f(m)$, but also of its Jacobian $\partial f / \partial m$. This finding provides empirical evidence that gradient information derived from neural networks can carry meaningful physical interpretation in geophysical inverse problems.

\subsection{Breakdown of Kernel Consistency at Long Periods}

Despite the high accuracy of forward predictions, the agreement between neural gradients and theoretical kernels deteriorates systematically at longer periods, particularly for Rayleigh waves. This discrepancy highlights a fundamental distinction between accurate forward modeling and recovery of physically consistent differential structure.

From a physical perspective, long-period surface waves have broader and smoother sensitivity kernels, which reflect reduced resolution and increased non-uniqueness in deeper regions. In such regimes, different subsurface models can produce nearly indistinguishable dispersion curves, leading to a poorly constrained inverse problem. As a result, the Jacobian of the forward operator becomes less well-defined, and small variations in the learned mapping can lead to large differences in gradient structure.

From a data-driven perspective, neural networks are trained to minimize prediction error in the observable space, rather than to preserve the exact differential properties of the underlying operator. When the mapping from model space to data space is weakly sensitive, the network may learn an effective representation that reproduces the input–output relationship but deviates from the true physical sensitivity. This effect is particularly evident in the appearance of spurious oscillations and incorrect sign structures in deep sensitivity regions.

\subsection{Differences Between Rayleigh and Love Waves}

The observed differences between Rayleigh and Love waves provide additional insight into the relationship between physical sensitivity and learned representations. Love-wave kernels remain comparatively more stable at long periods, whereas Rayleigh-wave kernels exhibit stronger degradation.

This contrast can be attributed to the different physical sensitivities of the two wave types. Love waves depend primarily on shear-wave velocity and exhibit simpler sensitivity structure, while Rayleigh waves are influenced by coupled variations in $V_s$, $V_p$, and density, leading to more complex and less well-conditioned sensitivity kernels. The increased complexity of Rayleigh-wave sensitivity may amplify the effects of non-uniqueness and make it more difficult for the neural surrogate to recover consistent differential structure at depth.

\subsection{Implications for Inversion and Interpretability}

These results have important implications for the use of neural network surrogates in geophysical inversion. The fact that neural gradients accurately recover shallow sensitivity structure suggests that they can be used as effective substitutes for theoretical kernels in gradient-based inversion, particularly in well-resolved regions. 

However, the degradation of kernel consistency at depth indicates that caution is required when interpreting neural gradients in poorly constrained regimes. In such cases, the gradients should be understood as effective sensitivities associated with the learned mapping, rather than exact representations of physical Fréchet kernels.

More broadly, our findings highlight a separation between forward accuracy and differential physical consistency in data-driven models. While neural networks can achieve near-perfect prediction performance, this does not guarantee that they preserve the full structure of the underlying physical operator. Understanding this distinction is critical for developing reliable and interpretable data-driven methods in geophysical applications. This behavior can also be interpreted in terms of information geometry, where regions of low sensitivity correspond to flat directions in the data misfit landscape, making the Jacobian inherently unstable and difficult to recover from data alone.

\subsection{Gradient-Based Inversion and Effective Jacobians}

\begin{figure}
    \centering
    \includegraphics[width=0.5\linewidth]{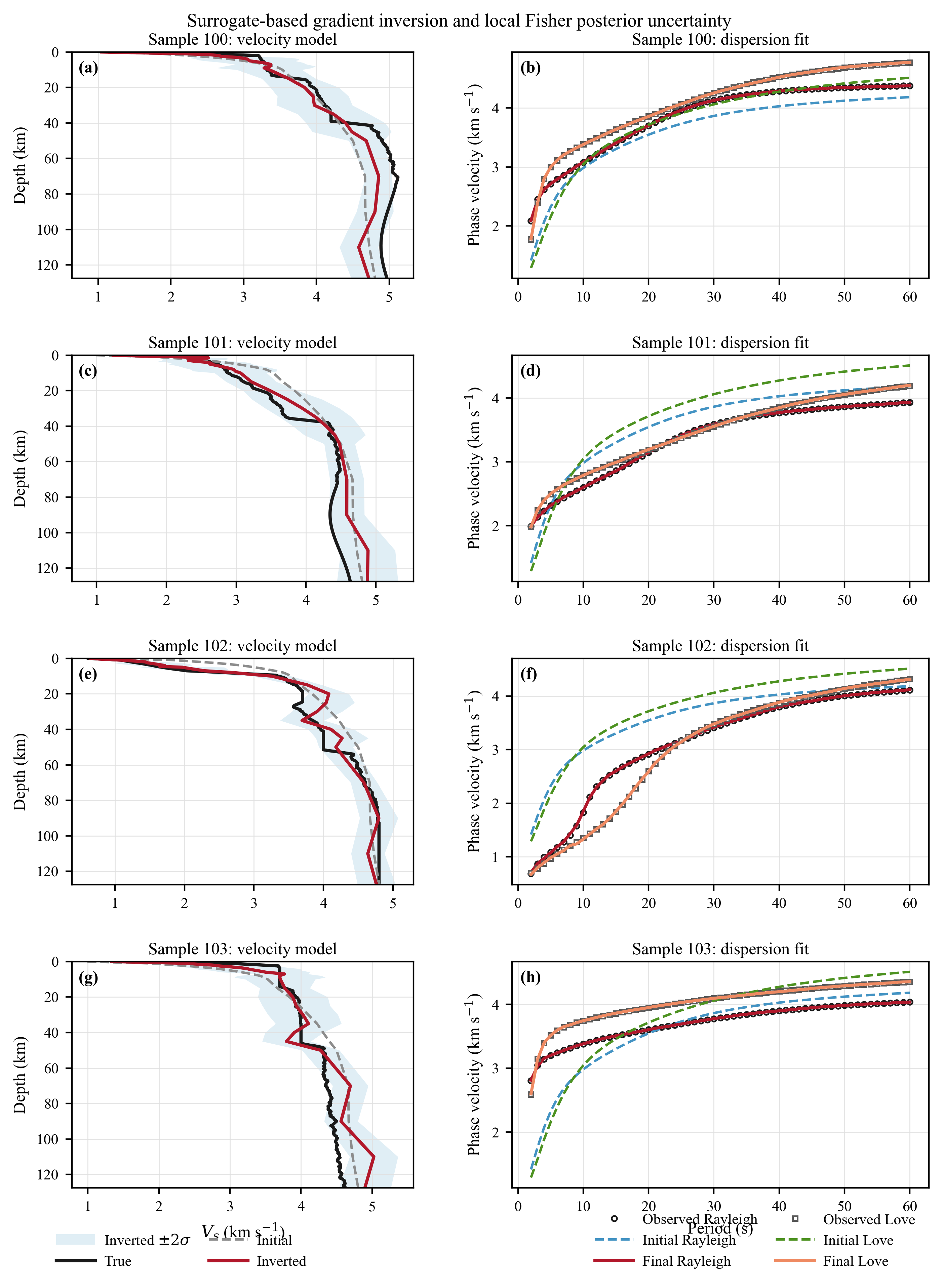}
\caption{
Surrogate-based gradient inversion and local posterior uncertainty estimated from the Fisher information matrix. Left panels show recovered shear-wave velocity models for four representative samples, including the true model (black), initial model (gray dashed), and inverted model (red). Shaded regions denote $\pm 2\sigma$ uncertainty derived from the inverse Fisher matrix. Right panels show corresponding dispersion fits for Rayleigh and Love waves before and after inversion.

The inversion successfully recovers the main structural features of the true model and achieves close agreement with observed dispersion curves. The posterior uncertainty increases with depth, reflecting reduced sensitivity and increased non-uniqueness in deeper regions. These results demonstrate that gradients derived from the neural surrogate provide a physically meaningful approximation of the forward operator Jacobian and can be used for both inversion and uncertainty quantification.
}
    \label{fig:fisher}
\end{figure}
To further assess the physical meaning of neural surrogate gradients, we examine their performance in gradient-based inversion and uncertainty quantification (Figure~\ref{fig:fisher}). 

Despite the deviations from theoretical sensitivity kernels observed at longer periods, the surrogate-derived gradients successfully guide the inversion toward velocity models that closely reproduce both the true structure and the observed dispersion curves. This indicates that the learned gradients provide a sufficiently accurate local linear approximation of the forward operator for optimization purposes.

The corresponding Fisher-based uncertainty estimates further support this interpretation. The inferred uncertainty exhibits a clear depth-dependent pattern, with low uncertainty in shallow regions and increasing uncertainty at depth, consistent with the decreasing sensitivity of long-period surface waves and the inherent non-uniqueness of the inverse problem. This agreement suggests that the neural surrogate captures the geometry of the inverse problem, even when its gradients deviate from theoretical kernels.

However, we find that the absolute amplitude of the Fisher-derived uncertainty requires empirical scaling to match the variability observed in the recovered models. This implies that while the surrogate preserves the relative curvature structure of the posterior, it does not fully recover its absolute scale. Such discrepancies likely arise from data normalization, implicit noise weighting, and the approximate nature of the Gauss--Newton/Fisher formulation.

Taken together, these results highlight an important distinction: neural surrogate models may not reproduce theoretical sensitivity kernels exactly, particularly in poorly resolved regions, but can still define effective Jacobians that preserve the functional geometry required for inversion and uncertainty analysis. This suggests that data-driven models can provide physically meaningful differential structure, even when deviating from classical kernel-based representations.

\subsection{Influence of Training Priors on Learned Sensitivities}

An additional feature observed in the surrogate-derived gradients is the presence of a persistent peak near $\sim70$ km at intermediate to long periods, which is not predicted by theoretical sensitivity kernels. Unlike physical kernels, whose peak sensitivity systematically shifts to greater depths with increasing period, this feature remains largely fixed in depth across different periods. Such period-independent behavior indicates that it does not originate from intrinsic wave propagation physics.

Controlled experiments further reveal that this anomaly is directly linked to the training model ensemble. When the surrogate is trained on models that systematically include upper-mantle low-velocity zones (LVZs) within a similar depth range, the $\sim70$ km peak consistently appears in the inferred gradients. In contrast, when the training ensemble is replaced by a weak-prior distribution without imposed structural features, this anomaly disappears and the gradients more closely follow theoretical sensitivity kernels. This provides direct evidence that the observed feature is a prior-induced artifact rather than a physically meaningful sensitivity.

These results demonstrate that neural network surrogates do not learn purely the physical sensitivity of the forward operator. Instead, the learned gradients reflect a combination of intrinsic physical sensitivities and statistical regularities embedded in the training data. In this sense, the surrogate gradients can be interpreted as an ``effective sensitivity'', shaped jointly by wave physics and prior information.

From an inverse-theory perspective, this implies that the Jacobian inferred from neural surrogates encodes not only the forward operator, but also the information geometry of the training distribution. As a result, gradient-based interpretations may be biased toward structures that are overrepresented in the prior model ensemble.

This finding has important implications for the use of data-driven forward models in inverse problems. While such models can provide useful and physically meaningful gradients, their interpretation requires careful consideration of the training distribution, particularly when strong structural priors are imposed. 

While Physics-Informed Neural Networks (PINNs) enforce physical consistency through explicit loss regularization, our results demonstrate that differential physical structures can emerge naturally from data alone, albeit with a heightened sensitivity to training priors.

\section{Conclusions}

In this study, we investigated whether data-driven forward models can recover the physical sensitivity kernels underlying surface-wave dispersion. Using a neural network surrogate trained to approximate the mapping between shear-wave velocity structure and dispersion observations, we analyzed gradients obtained via automatic differentiation and compared them with theoretical Fréchet kernels.

Our results show that neural surrogate models are able to recover the dominant depth-dependent sensitivity structure of surface waves, particularly at short and intermediate periods. In these regimes, the gradients closely match theoretical kernels in both shape and depth localization, and can be directly used to guide gradient-based inversion. Furthermore, Fisher-based uncertainty estimates derived from these gradients exhibit physically meaningful depth-dependent behavior, consistent with classical resolution analysis.

However, systematic deviations emerge at longer periods and greater depths, especially for Rayleigh waves. While the forward predictions remain highly accurate, the corresponding gradients increasingly diverge from theoretical sensitivity kernels, reflecting the reduced resolution and increased non-uniqueness of the inverse problem in these regimes. This indicates that accurate forward modeling does not necessarily imply full recovery of the differential physical structure of the underlying operator.

These findings suggest that data-driven forward models can provide an effective and physically interpretable approximation of both the forward mapping and its associated sensitivity structure in well-resolved regions, while deviations at depth highlight fundamental limitations imposed by information content and model non-uniqueness. More broadly, this work demonstrates that the differential structure of geophysical forward operators can partially emerge in data-driven models, providing a pathway toward integrating machine learning with physics-based inversion and uncertainty quantification frameworks.

\section*{Open Research}
The code used to generate the results in this study is publicly available at \url{https://github.com/cangyeone/physical-sensitivity-kernels.git}.

\section*{Conflict of Interest}
The authors declare no conflicts of interest relevant to this study.

%
%


%
%
%
%
%
\bibliography{agusample}

\end{document}